\documentclass[manuscript]{acmart}

\usepackage{amsmath,amsfonts}
\usepackage{algorithmic}
\usepackage{booktabs}
\usepackage{graphicx}
\usepackage{hyperref}
\usepackage{multirow}
\usepackage{multicol}
\usepackage{subfigure}
\usepackage{textcomp}
\usepackage{xcolor}

\hypersetup{
    colorlinks=true,
    linkcolor=blue,
    filecolor=magenta,      
    urlcolor=cyan,
    pdftitle={Overleaf Example},
    pdfpagemode=FullScreen,
    }

\def\BibTeX{{\rm B\kern-.05em{\sc i\kern-.025em b}\kern-.08em
    T\kern-.1667em\lower.7ex\hbox{E}\kern-.125emX}}

\AtBeginDocument{%
  \providecommand\BibTeX{{%
    \normalfont B\kern-0.5em{\scshape i\kern-0.25em b}\kern-0.8em\TeX}}}

\copyrightyear{2023}
\acmYear{2023}
\setcopyright{licensedothergov}\acmConference[IoTDI '23]{International Conference on Internet-of-Things Design and Implementation}{May 9--12, 2023}{San Antonio, TX, USA}
\acmBooktitle{International Conference on Internet-of-Things Design and Implementation (IoTDI '23), May 9--12, 2023, San Antonio, TX, USA}
\acmPrice{15.00}
\acmDOI{10.1145/3576842.3582362}
\acmISBN{979-8-4007-0037-8/23/05}

\begin{document}

\title{Because Every Sensor Is Unique, so Is Every Pair: Handling Dynamicity in Traffic Forecasting}


\author{Arian Prabowo}
\email{arian.prabowo@gmail.com}
\orcid{0000-0002-0459-354X}
\affiliation{
  \institution{RMIT University}
  \city{Melbourne}
  \state{Victoria}
  \country{Australia}
}
\affiliation{
  \institution{Data61/CSIRO}
  \city{Melbourne}
  \state{Victoria}
  \country{Australia}
}

\author{Wei Shao}
\email{weishao@ucdavis.edu}
\orcid{0000-0002-9873-8331}
\affiliation{
  \institution{UC Davis}
  \city{Davis}
  \state{California}
  \country{USA}
}

\author{Hao Xue}
\email{hao.xue1@unsw.edu.au}
\orcid{0000-0003-1700-9215}
\affiliation{
  \institution{UNSW}
  \city{Sydney}
  \state{New South Wales}
  \country{Australia}
}

\author{Piotr Koniusz}
\email{piotr.koniusz@anu.edu.au}
\orcid{0000-0002-6340-5289}
\affiliation{
  \institution{Data61/CSIRO}
  \city{Canberra}
  \state{Australian Capital Territory}
  \country{Australia}
}
\affiliation{
  \institution{ANU}
  \city{Canberra}
  \state{Australian Capital Territory}
  \country{Australia}
}

\author{Flora D. Salim}
\email{flora.salim@unsw.edu.au}
\orcid{0000-0002-1237-1664}
\affiliation{
  \institution{UNSW}
  \city{Sydney}
  \state{New South Wales}
  \country{Australia}
}


\begin{abstract}
Traffic forecasting is a critical task to extract values from cyber-physical infrastructures, which is the backbone of smart transportation. However
% , due to its complex non-linear dynamics, traffic forecasting is challenging. The traffic dynamics consist of many relationships, such as the temporal ones between the past and future speed, the spatial ones in terms of the speed recorded at two different sensors, and the ones between the speed and volume. Moreover, 
owing to external contexts, the dynamics at each sensor are unique. For example, the afternoon peaks at sensors near schools are more likely to occur earlier than those near residential areas.
In this paper, we first analyze real-world traffic data to show that each sensor has a unique dynamic. Further analysis also shows that each pair of sensors also has a unique dynamic. Then, we explore how node embedding learns the unique dynamics at every sensor location. Next, we propose a novel module called Spatial Graph Transformers (SGT) where we use node embedding to leverage the self-attention mechanism to ensure that the information flow between two sensors is adaptive with respect to the unique dynamic of each pair. Finally, we present Graph Self-attention WaveNet (G-SWaN) to address the complex, non-linear spatiotemporal traffic dynamics. Through empirical experiments on four real-world, open datasets, we show that the proposed method achieves superior performance on both traffic speed and flow forecasting.
Code is available at: \url{https://github.com/aprbw/G-SWaN}
\end{abstract}




\begin{teaserfigure}
     \centering
     \subfigure[Locations of the sensors on the Californian highway network surrounding the bay area. Installing a network of sensors on a road infrastructure enables traffic forecasting and smarter cities. \label{fig:tf1}]{
         \includegraphics[width=.45\columnwidth]{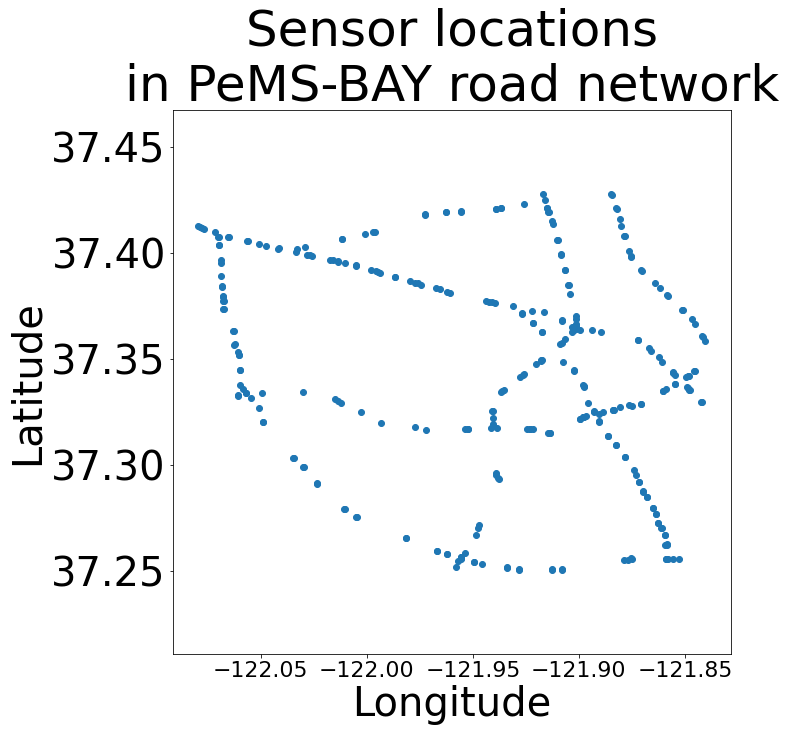}
     }
     \hspace{1cm}
     \subfigure[At each sensor, traffic forecasting uses the recent sensor readings (solid blue line) to predict the future traffic (red line). This forecast is made by our proposed architecture Graph Self-attention WaveNet (G-SWaN). Our forecasts accurately predict the future traffic (dotted blue line). \label{fig:tf2}]{
         \includegraphics[width=.4\columnwidth]{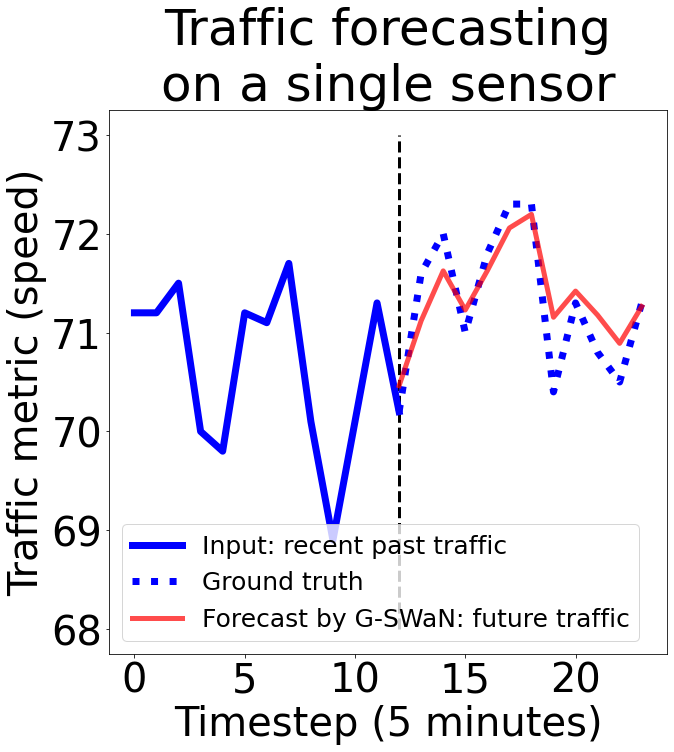}
     }
    \caption{Visual abstract of the traffic forecasting task.}
    \label{fig:tf}
\end{teaserfigure}

\maketitle

\section{Introduction}

The recent proliferation of networked physical sensors have allowed exponential growth of machine generated data in many applications, including road traffic. This has led to the emergence of intelligent transportation systems, which have the potential to revolutionize the way we live.

% However, owing to its complex non-liner dynamics, traffic data is challenging to learn from in huge volumes. The traffic dynamics consist of many relationships, such as the temporal ones between the past and future speed, the spatial ones in terms of the speed recorded at two different sensors, and the ones between the speed and volume. Therefore, we ask the following questions: 1) How can we learn the dynamics of each sensor in a data-driven manner? 2) How can we learn the dynamics between pairs of sensors when the underlying connections between sensors are not given? 3) How can we jointly learn both dynamics end-to-end to generate an effective traffic forecasting?

Previous works have discussed and explored the challenges of traffic forecasting (Figure \ref{fig:tf}) relating to the complexity of temporal and spatial dynamics.
Temporal dynamics is the relationships between past and future traffic.
Although many simplified mathematical model exists \cite{bellomo2011modeling}, none are capable to capture the complexity of real world traffic.
This resulted in many research towards data-driven methods \cite{zhang2018gaan,yu2017stgcn,cho2014learning,huang2019dsanet,guo2019attention,song2020spatial,wu2020graph}.
Meanwhile, spatial dynamics is the relationships traffic at different locations.
The complexity in spatial dynamics is caused by the non-euclidean topology of the road network \cite{li2017diffusion,yu2017stgcn,guo2019attention,bai2020adaptive, wu2020graph, song2020spatial}.

% To begin, consider the variables involved in traffic forecasting.
% Traffic forecasting (Figure \ref{fig:tf}) predicts the future traffic metrics (e.g., speed and flow) at multiple sensors in a road network.
% This is challenging because the temporal dynamics (i.e., between past and future traffic) are complex and non-linear \cite{zhang2018gaan,yu2017stgcn,cho2014learning,huang2019dsanet,guo2019attention,song2020spatial,wu2020graph}.
% Second, the spatial dynamics of the sensors are non-euclidean as they are dependent on the topology of the road network \cite{li2017diffusion,yu2017stgcn,guo2019attention,bai2020adaptive, wu2020graph, song2020spatial}.

\begin{figure}[htbp]
    \centering
    \subfigure[Idealized fundamental diagram. Figure is copied from \cite{bellomo2011modeling}. \label{fig:fund_MAR}]{
        \includegraphics[width=.35\columnwidth]{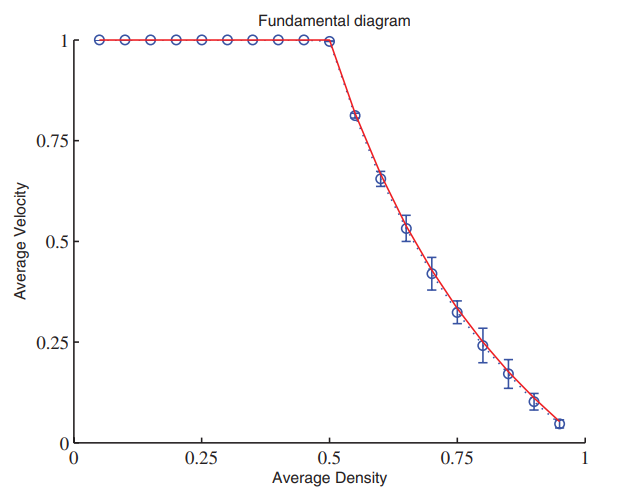}
    }
    \hfill
    \subfigure[Fundamental diagram of a sensor in PeMS-D4 dataset. \label{fig:fund1}]{
        \includegraphics[width=.25\columnwidth]{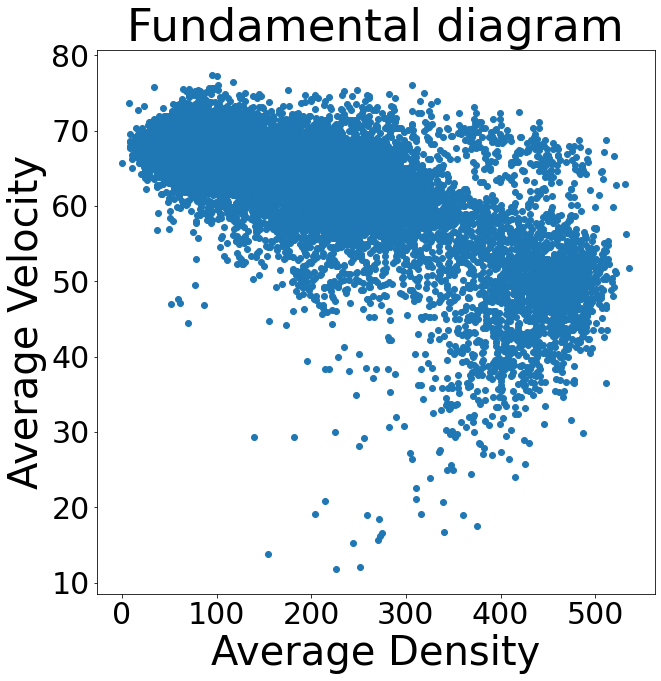}
    }
    \hfill
    \subfigure[Fundamental diagrams of 12 selected sensors in PeMS-D4 dataset showing great diversity. \label{fig:fund_many}]{
        \includegraphics[width=.35\columnwidth]{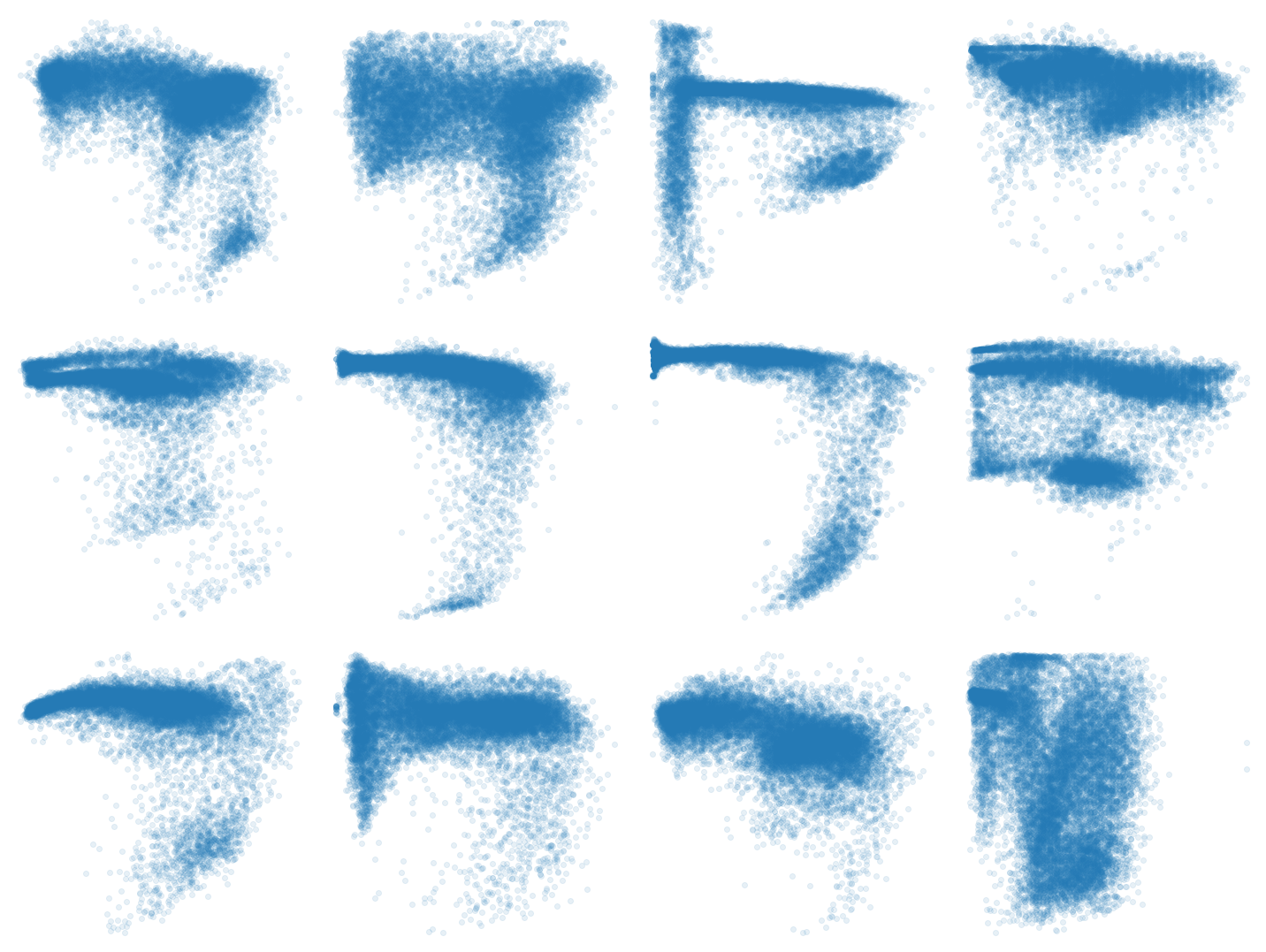}
    }
    \caption{Fundamental diagrams showing the dynamics between flow (density) in the x-axis and speed (velocity) in the y-axis. Figure \ref{fig:fund_MAR} and \ref{fig:fund1} compare the idealized fundamental diagram against the real-world data, while Figure \ref{fig:fund_many} shows that every sensor has a unique fundamental diagram, suggesting a unique underlying dynamics. The PeMS-D4 dataset is detailed in section \ref{sec:dataset}. \label{fig:fund}}
\end{figure}

% Third, every sensor has a unique dynamic.
However, one novel observation presented in this paper is that every sensor has a unique dynamics.
We show this by plotting flow (density) versus speed (velocity), also known as fundamental diagrams \cite{bellomo2011modeling}.
Figure \ref{fig:fund_MAR} shows the idealized fundamental diagram based on a mathematical model of traffic \cite{berthelin2008model}.
According to this model, all sensors have the same general shape: The speed stays constant during a free flow of traffic and decreases during a congested flow.
The real-world data, however, reveals a different dynamic.
Figure \ref{fig:fund1} shows the fundamental diagram of a sensor in the PeMS-D4 dataset.
Although both the theory and the data agree that there is an inverse relationship between flow and speed, the sharp threshold between free flow and congested flow is not found in this particular sensor data.
At every density, moreover, there is a huge variance of speed. 
Finally, Figure \ref{fig:fund_many} shows that there is a great diversity between fundamental diagrams of various sensors, strongly suggesting that every sensor has a unique dynamic.
This suggests the need of a mechanism to capture individual sensor dynamics.

\begin{figure}[htbp]
    \centering
    \hfill
    \subfigure[Association plots of 8 random sensor pairs. \label{fig:pairs}]{
        \includegraphics[width=.6\columnwidth]{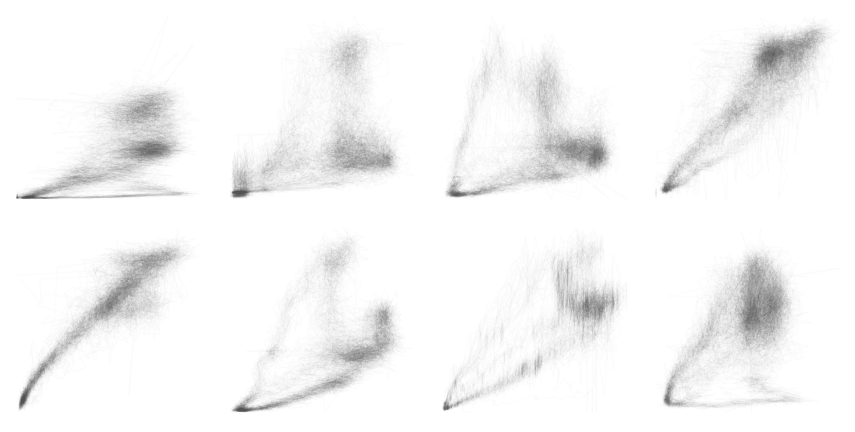}
    }
    \hfill
    \subfigure[Association plots of a sensor pair over three weeks. Blue lines are weekdays, red lines are weekends. \label{fig:pair_over_time}]{
        \includegraphics[width=.35\columnwidth]{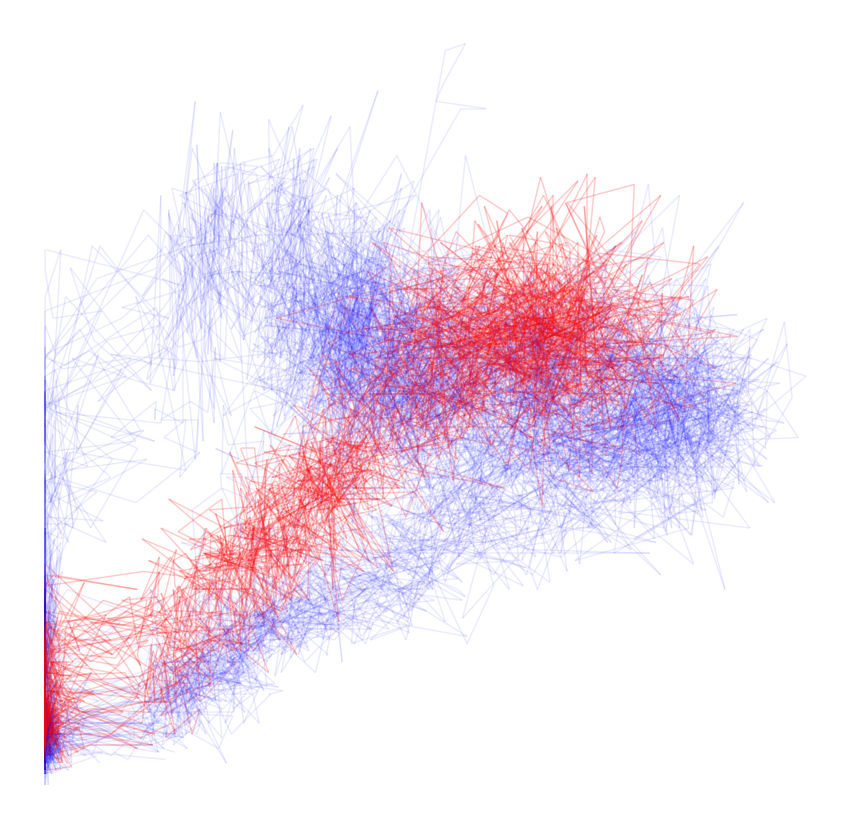}
    }
    \hfill \null
    \caption{Association plots of different pairs of sensor readings in PeMS-D4. The x-value of a point is the flow at one sensor, while the y-value is the flow at the other sensor. Consecutive data points are connected by a line. The PeMS-D4 dataset is detailed in section \ref{sec:dataset}. \label{fig:pair}}
\end{figure}

Furthermore, given that all individual sensor dynamics are unique, the impact traffic at a sensor location has on its neighbors also involves complex relationships that are unique to that pair.
The data confirms this behavior.
In Figure \ref{fig:pairs}, we plot the flow correlation of several pairs of sensors.
The plot shows great diversity in the dynamics of sensor pairs.
Moreover, Figure \ref{fig:pair_over_time} shows that even within a single sensor, the dynamic changed periodically over time.
The weekday pair dynamic (blue) is distinct from the weekend pair dynamic (red).

% more examples or motivating scenario.
There are many other ways to formulate this task besides as a multivariate timeseries forecasting, one example being as conditional spatiotemporal generation \cite{gao2020generative}.
Another is a mathematical model of traffic, which can explain the cause of unique sensor and sensor pair dynamics.
The macroscopic or continuum model of traffic is based on partial differential equations (PDEs) \cite{van2015TrafficGenealogy}.
In this model, traffic is treated as a continuous fluid, with aggregate values used such as average speed and flow.
Similar to fluid dynamics, traffic is modelled using a set of PDEs and boundary conditions.
While initial conditions are often discussed in the context of PDEs, they have less relevance in this context and will be ignored.

The PDEs that describe the evolution of traffic flow have spatial shifts or translation symmetry, meaning that they are the same everywhere.
This is the dynamics that is being captured by most existing models that does not take sensor and sensor pairs unique dynamics into account.
However, different places may have different boundary conditions, breaking the spatial symmetry and resulting in unique dynamics for every sensor.
For example, entry and exit ramps can be modelled as inlet/source and outlet/sink boundary conditions, respectively;
lane or road closures can be considered changes to the geometry of the boundaries.
This is why traffic flow may differ significantly between neighbouring sensors, even if they are connected by a road.

To address these challenges, we present Graph Self-attention WaveNet (G-SWaN).
It contains a novel spatial module called Spatial Graph Transformers (SGT) that extends a self-attention mechanism \cite{vaswani2017attention} to traffic networks with unique sensor dynamics.
Moreover, being a graph neural network, SGT are well suited to deal with the non-euclidean spatial dynamics of a traffic network \cite{bronstein2021geometric}.
The node embedding in SGT injects sensor identity information into the self-attention mechanisms, enabling them to adapt to the unique dynamics of each pair.

To show the generalizability of our architecture in traffic forecasting tasks, we evaluate our model by two traffic metrics: Speed and flow.
We use four public, real-world datasets for reproducibility.
Briefly, our main contributions are as follows:

\begin{enumerate}
    \item To adapt to the behavior dynamics of individual sensors and pairs of sensors, as shown in Figure \ref{fig:fund} and \ref{fig:pair}, we propose the Spatial Graph Transformers (SGT) module, which encodes individual sensor dynamics through node embedding.
    \item We introduce a novel traffic forecasting architecture, Graph Self-attention WaveNet (G-SWaN), which uses SGT to capture the unique dynamics.
    \item Through extensive experiments on four open real-world datasets, we empirically show that the proposed architecture achieves state-of-the-art results.
    
\end{enumerate}

\section{Related Work} 

This section discusses existing works on traffic forecasting, starting from classical statistical techniques to the latest deep learning methods, including attention-based models.

One of the first works in this area \cite{ahmed1979analysis} featured the Box-Jenkins technique. A large number of these earlier works also fall into the data-driven and statistical approaches to machine learning. These include Autoregressive Integrated Moving Average (ARIMA) \cite{hamed1995short} and ARIMA-like approaches, such as KARIMA \cite{van1996combining}, subset ARIMA \cite{lee1999application}, ARIMAX \cite{williams2001multivariate}, VARMA \cite{kamarianakis2003forecasting}, and SARIMA \cite{williams2003modeling}, as well as classical machine learning methods like SVR \cite{jeong2013supervised,lippi2013short,chen2012retrieval}.

Based on early successes in various other tasks, deep learning was applied to traffic forecasting and \cite{lv2014traffic} used stacked autoencoder as a generic latent feature learning.
This was followed by a number of works that used varieties of Recurrent Neural Networks (RNN), such as Gated Recurrent Units (GRU) \cite{fu2016using} and Long Short-Term Memory (LSTM) \cite{shao2020deep}\cite{cui2020graph}, to better capture the temporal dynamics.

In contrast to the temporal focus used in the aforementioned studies, other works captured spatial information from traffic data to infer road maps \cite{prabowo2019coltrane} and predict flight delays \cite{shao2019flight,shao2022predicting}.
Diffusion Convolution Recurrent Neural Network (DCRNN) \cite{li2017diffusion} attempted to capture both spatial and temporal dynamics by alternating between spatial and temporal modules.
It established the architecture for many subsequent works, including the present one.

DCRNN also extended convolution from undirected to directed graphs, arguing that traffic flow is directed in a network.
In addition, Spatio-Temporal Hashing Multi-Graph Convolutional Network (ST-HMGCN) extended convolution to a multi-graph in order to forecast in a bus transit system, arguing that there are many different types of dependencies beyond the spatiotemporal ones, such as the semantics of different stations \cite{luo2021sthmgcn}.

\begin{table}[htbp]
\normalsize
\caption{Notations}\label{tab:notation}
\centering
\begin{tabular}{@{}cl@{}}
\toprule
Notation & Description \\ \midrule

$\alpha$ & attention matrices \\
$A$ & adjacency matrices \\
$A_{adp}$ & adaptive adjacency matrices \\
[4pt]

$A_r$ & \begin{tabular}[c]{@{}l@{}}adjacency matrices based on \\ physical road network connectivity\end{tabular} \\
[9pt]

{[}B,D,N,L{]} & \begin{tabular}[c]{@{}l@{}} tensor shape: \\ B: batch size, \\ D: embedding dimension size, \\ N: number of sensors,\\ L: number of timesteps in an observation window \end{tabular} \\
[27pt]

$D_r$ & final representation dimension size \\
$\mathbf{e}_1$, $\mathbf{e}_2$ & node embeddings \\
$F$ & forecasting horizon in timesteps \\
FC & Fully Connected layer \\
GCN & Graph Convolutional Networks \\
$H$ & number of attention heads \\
$\mathbf{h}$ & model output, prediction on the future traffic \\
MAE & Mean Absolute Error \\
MAPE & Mean Percentage Error \\
[4pt]

\begin{tabular}[c]{@{}l@{}} query ($Q$) \\ key ($K$) \end{tabular} & \begin{tabular}[c]{@{}l@{}} query and key matrices as a part of \\ self-attention mechanisms \end{tabular} \\
[11pt]

RBF & Radial Basis Functions \\
RMSE & Root Mean Square Error \\
$\sigma$ & sigmoid activation function \\
$tanh$ & hyperbolic tan \\
tod & time-of-day \\
$W$ & number of G-SWaN layers \\
$\mathbf{x}$ & feature data point \\
$\mathbf{y}$ & ground truth label \\

\bottomrule
\end{tabular}
\end{table}

Following this general approach, Spatio-Temporal Graph Convolution Network (STGCN) \cite{yu2017stgcn} replaces the GRU in DCRNN with 1D convolution layers in order to increase efficiency.
This pattern of replacing a specific module with inspiration from other tasks continues with Graph WaveNet \cite{wu2020graph,shleifer2019incrementally}.
By replacing temporal 1D convolution with WaveNet, the study \cite{oord2016wavenet}, showed success in capturing temporal dynamics for audio.
They also replaced the spectral GCN \cite{kipf2016semi,zhu2021contrastive,zhu2022generalized} with spatial GCN \cite{hamilton2017inductive}.
The limitation of the adjacency matrix is that it only reflects the physical connectivity between sensors to capture the latent spatio-temporal correlations \cite{de2020inclusion}.
This line of reasoning motivates the development of adaptive graph generation \cite{wu2020graph,wang2020traffic,bai2020adaptive}.

Some later methods took inspiration from the successes of attention mechanisms \cite{bahdanau2014neural} applied to spatial GCN \cite{velickovic2017graph}, combined with the self-attention paradigm of transformer \cite{vaswani2017attention,zhang2018gaan,brown2020language,chen2020generative,wang2020traffic}, and introduced their own variants of GAT such as \cite{park2019stgrat,zheng2020gman,kong2020stgat,lu2020spatiotemporal}.
This contrasts with the existing attention-based works on traffic forecasting, such as Attention Based Spatial-Temporal Graph Convolutional Network (ASTGCN) \cite{guo2019attention} and Reinforced Spatiotemporal Attentive Graph Neural Networks (RSTAG) \cite{zhou2020rstag}, which only focus on past data instead of the road network structure.
Attentive Spatial-Temporal Convolutional Network (ASTCN) has also used an attention mechanism to forecast traffic flow on a grid instead of on a graph \cite{guo2021astcn}.

Attention mechanisms have retained their popularity over recent years \cite{xue2021mobtcast,xue2021termcast}.
One approach \cite{li2021spatiotemporal} used GAT to forecast traffic metrics of the edge attributes rather than the sensor attributes, while another \cite{fang2021ftpg} used an attention-based method on GPS traces to get a fine-grained representation.
More closely related to our present work, other works \cite{abdelraouf2021utilizing, zhang2021graph} used attention to capture temporal dynamics.
% However, while the first work focused on a road segment instead of a network, the latter work constructed a static latent adjacency matrix based on the physical distances between sensors.
% In contrast, we use spatial attention to make the adjacency matrices adaptive.
Nevertheless, none of these works took into account the node and edges unique dynamics, nor provided a method to effectively capture these dynamics.

\section{Method}
\subsection{Problem Statement}

The dataset
$\mathbf{X} \in \mathbb{R}^{D_{input} \times N \times K}$
is a tensor where
$d \in D_{input}$ includes the traffic measurements and other contextual information,
such that $d=0$ is the traffic measurement
and $d=1$ is the $k^{th}$ the time-of-day,
$N$ is the number of recording stations,
and
$K$ is the number of timesteps.
Each data point
$\mathbf{x}_k = \mathbf{X}_{D_{input},N,k:k+L}$
is a tensor, where $L$ is the number of timesteps in the data point.

Road networks are abstracted to a directed graph with weighted edges. This graph is represented through a sparse adjacency matrix $A_r$. The adjacency matrix is normalized between zero and one. An edge with higher weight means that the sensors are closer together, while lower weight means that the sensors are further apart.

The traffic forecasting task is a multi-step forecasting problem formalized as follows:
$\mathbf{h}(\mathbf{x}_k)=\mathbf{x}_{k+L+F}$
where $F$ is the forecasting horizon.
A description of the notation used is presented in Table \ref{tab:notation}.

\subsection{Graph Self-attention WaveNet (G-SWaN)}

\begin{figure*}[htbp]
\centering
\includegraphics[width=\textwidth]{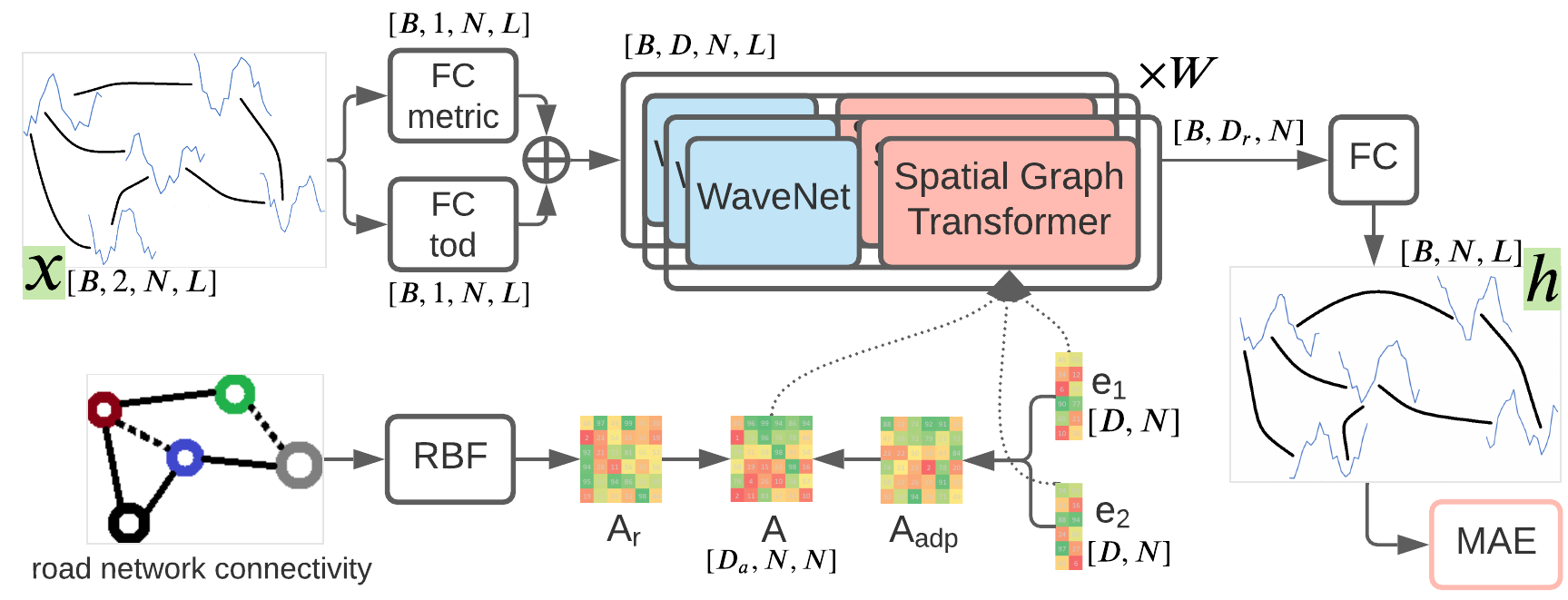}
\caption{System architecture of Graph Self-attention WaveNet (G-SWaN). Spatial Graph Transformers (SGT) is the novel module proposed that uses the node embeddings $e_1$ and $e_2$ to capture the unique sensor dynamics in the self-attention mechanisms. The notations are described in Table \ref{tab:notation}.}
\label{fig:sysarc}
\end{figure*}

The general structure of the proposed Graph Self-attention WaveNet (G-SWaN) is shown in Figure \ref{fig:sysarc}.
While we use Graph WaveNet \cite{wu2020graph} as our backbone, we also make a number of minor improvements.
The main contribution in this paper is the SGT module.

The major components of G-SWaN are augmentations, initial embedding, G-SWaN layers, and the final fully connected (FC) layers.
The initial embedding module, which functions as the encoder for the input $x$, consists of two parallel FC layers.
The output is then passed through $W$ G-SWaN layers, each consisting of a WaveNet module to learn the temporal dynamics and a SGT module to learn the spatial dynamics.

The SGT uses adjacency matrices $A$ and node embeddings $e_1$ and $e_2$.
These node embeddings learn the node and edge unique dynamics end-to-end.
The adjacency matrices $A$ are composed of an adjacency matrix based on the physical connectivity of the road network $A_r$ and a learned adaptive one $A_{adp}$.
$A_r$ is constructed by passing the Euclidean distance of physically connected sensors through a Radial Basis Function (RBF).
Meanwhile, $A_{adp}$ is constructed as a batched dot product of the node embeddings.
Most importantly, SGT uses a self-attention mechanism to make the adjacency matrices $A$ adaptive to the current input.
Finally, the output of the G-SWaN layers is passed through FC layers to produce the forecasted future traffic $h$.

The loss function is Mean Absolute Error (MAE).
In the tensor shapes,
B is the batch size,
D is the embedding dimension,
N is the number of sensors, and
L is the number of timesteps.

Many minor elements such as augmentation \cite{zhang2023spectral}, activation functions \cite{zhang2022GReLU}, batch normalization, tensor reshape, and batch operations are not shown.
The major components introduced in this section will be described in the order above.

\subsubsection{Augmentation} \label{sec:aug}

We use three different augmentations:
\begin{itemize}
    \item \textbf{Soft spatial occlusion}. For each data point, every station has a 5\% probability to be partially occluded through a 0.05 scaling.
    \item \textbf{Temporal permutation}. For each data point, every timestep has a 5\% probability to have the stations be permuted.
    \item \textbf{Uniform noise}. A uniform noise with a scale of 5\% of the standard deviation is added to each entry of a data point.
\end{itemize}

\subsubsection{Initial Embedding}
The initial embedding consists of two parallel FC layers.
Functioning as an encoder, both project the input to the latent space with an embedding size of $D$.
The first layer (FC metric) projects the historical traffic data, while the second layer (FC tod) projects the time-of-day (tod) information.
Because both projections have the same dimensions, they can be aggregated via summation.

\subsubsection{G-SWaN Layer\label{sec:gwn}}

A G-SWaN layer is a spatio-temporal layer made up of three main components: WaveNet \cite{oord2016wavenet} as the temporal module, SGT as the spatial module, and an FC layer.
The WaveNet module takes the outputs of the previous layer as the input and outputs to the SGT module.
The SGT module outputs to the next G-SWaN layer.
In the earlier layers, SGT is building simpler spatial latent features as it only has a small temporal receptive field.
However, in the later layers, SGT is building a richer spatiotemporal latent feature as the receptive field covers the entire observation window.

To deal with diminishing gradients, the outputs of both the temporal and spatial module are also connected to the final FC layer in the G-SWaN layer to form residual connections \cite{he2016deep}.
This FC layer has $D_r$ neurons.
The outputs of each FC layer is then aggregated via summation.
This is detailed in Figure \ref{fig:G-SWaN}.
The final FC are the ones below the gray area.
The WaveNet and SGT modules are described in the following two sections.

\subsubsection{WaveNet}

WaveNet \cite{oord2016wavenet} is a convolutional alternative to RNN that deals with sequential data.
By using dilation, it gives exponential increases of receptive field width with respect to the number of layers, as opposed to traditional CNN.

Following the original Graph WaveNet \cite{wu2020graph}, two WaveNet are used per layer in parallel. One is activated with $tanh$ and acts as a convolutional filter, while the other is activated with sigmoid and acts as a gating mechanism on each latent channel.
The outputs of the WaveNet module are passed on to the SGT.
Additionally, a residual connection is passed to the final FC layer of the corresponding G-SWaN layer.

\subsubsection{Spatial Graph Transformer (SGT)}
\label{sec:sgt}

\begin{figure*}[htbp]
\centering
\includegraphics[width=\textwidth]{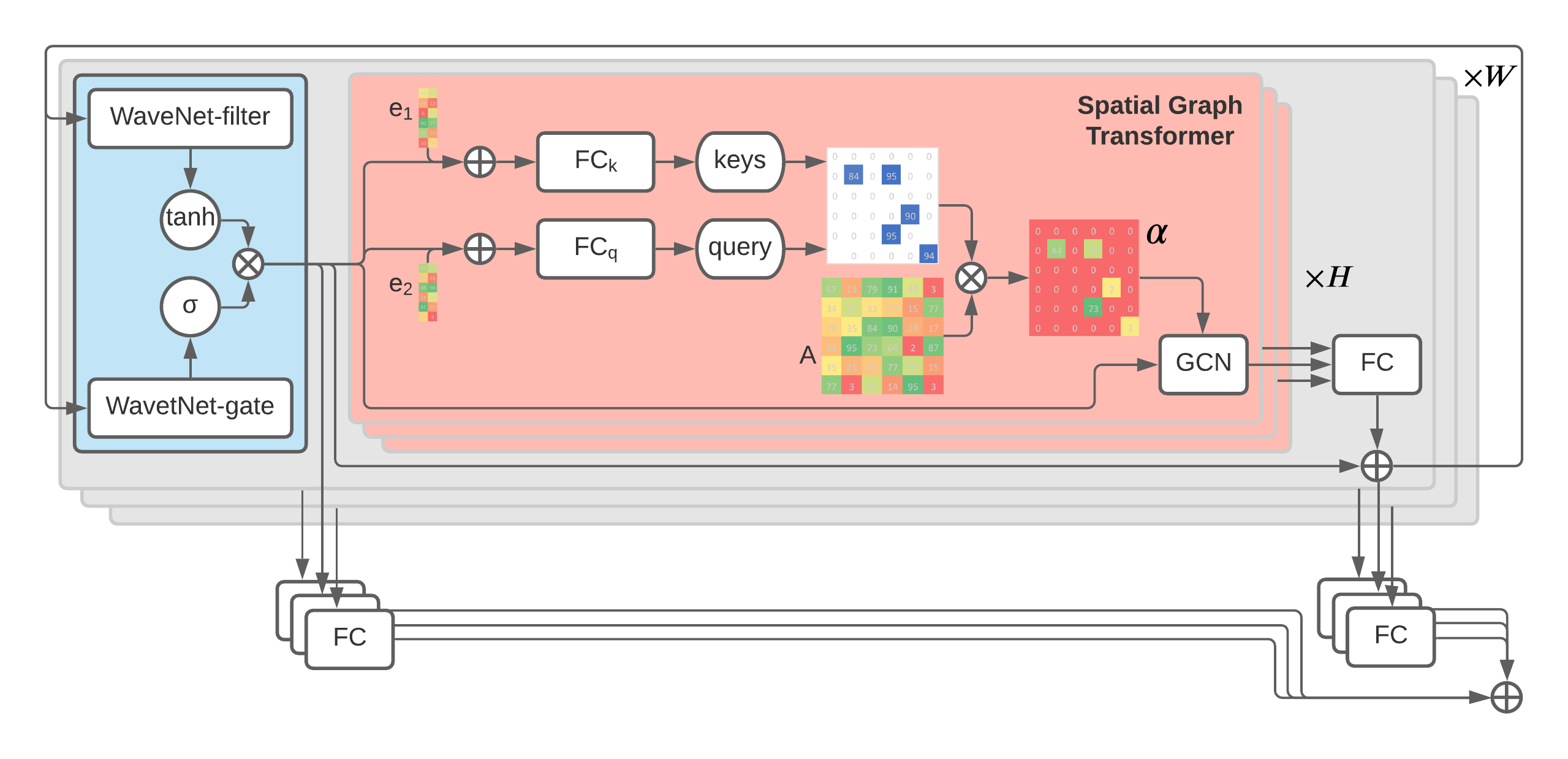}
\caption{
A layer of G-SWaN contains a WaveNet temporal block (blue) and a Spatial Graph Transformer (SGT) block (red).
SGT uses node embedding ($e_1$ and $e_2$) to apply multi-headed a query and key self-attention mechanism on the adjacency matrix $A$.
This way, the self-attention mechanism is sensitive to the unique dynamics of every pair of sensors.
There are $H$ attention heads and $W$ layers.
Some details such as activation functions and
batch normalization are also not shown.
}
\label{fig:G-SWaN}
\end{figure*}

When modelling the spatial dynamics, it is important to factor in that spatial dynamics change periodically through time, as shown in Figure \ref{fig:pair_over_time}.
However, many current works do not take this into account.
Here are three examples from STGCN \cite{yu2017stgcn}, AGCRN \cite{bai2020adaptive}, and Graph WaveNet \cite{wu2020graph}, respectively:
\begin{align}
    \mathbf{x}' &= \mathbf{W}(\mathbf{A}_L) \mathbf{x} ,\\
    \mathbf{x}' &= \mathbf{W}_E \mathbf{W}_W \mathbf{x} \mathbf{A}_L ,\\
    \mathbf{x}' &= \sum^K_k
    \mathbf{W}_{1,k} \mathbf{x} \mathbf{A}_p^k +
    \mathbf{W}_{2,k} \mathbf{x} \mathbf{A}_{adp}^k ,
\end{align}
where $\mathbf{x}$ and $\mathbf{x}'$ are the input and output from and to the previous and subsequent layers, respectively,
$\mathbf{W}$, $\mathbf{W}_E$, $\mathbf{W}_W$, $\mathbf{W}_{1,k}$, and $\mathbf{W}_{1,k}$ are all learnable parameters,
$\mathbf{A}_L$ is the Laplacian of the adjacency matrix \cite{kipf2016semi},
$\mathbf{A}_p$ is the power normalized adjacency matrix \cite{li2017diffusion},
$\mathbf{A}_{adp}$ is the adaptive adjacency matrix,
and
$K$ is the number hops in the spatial diffusion.

In all three cases, the adjacency matrices $\mathbf{A}$ are not a function of the input signal $\mathbf{x}$, rendering them incapable to adapt to the changes in the spatial dynamics.
To address this issue, we propose a novel module called SGT to make the GCN adaptive to the unique and changing dynamics of each pair of sensors.
This has some similarity with the attention mechanism in GAT \cite{velickovic2017graph}.
SGT uses a transformer-like query, key, and value formulation of self-attention \cite{vaswani2017attention}.

We propose the following graph convolution layer:
\begin{align} \begin{split}
    \mathbf{x}'(\mathbf{x}|\mathbf{e_1},\mathbf{e_2}) = 
\sum_{k,h}^{K,H} \mathbf{W}_{1,k} \mathbf{x}
\alpha(\mathbf{A}_r|\mathbf{x},\mathbf{e_1},\mathbf{e_2})^k
+ \mathbf{W}_{2,k} \mathbf{x} \alpha(\mathbf{A}_{adp}|\mathbf{x},\mathbf{e_1},\mathbf{e_2})^k
\end{split} \end{align}

where
$\mathbf{e_1}$ and $\mathbf{e_2}$ are the source and target node embeddings,
$H$ is the number of attention heads,
$\mathbf{A}_r$ is the adjacency matrix based on the physical road connectivity,
and finally the $\alpha(\mathbf{A}|\mathbf{X},\mathbf{e}_1,\mathbf{e}_2)$ function is the attention mechanism that is adaptive to the input signal $x$, which is going to be described in more detail in the following paragraph.
The $\sum$ operation could be generalized to any aggregating function.
In our case, $\sum$ is implemented as two FC layers, shown in Figure \ref{fig:G-SWaN} to the right of the red area and at the bottom right of the grey area.

$\mathbf{A}_{adp}$ is a learned adjacency matrix to capture complex spatial relations that is not captured through the former adjacency matrix ($\mathbf{A}_r$).
This is shown in the bottom row of Figure \ref{fig:sysarc}.
The adaptive adjacency matrix is calculated as follows:
\begin{equation}
    \mathbf{A}_{adp} =
SoftMax(ReLU(\mathbf{e_1}\mathbf{e_2}^T)).
\end{equation}

Finally, $\alpha(\cdot)$ is a self-attention mechanism that dynamically determines the attention given to each edge, based on the current input and node embeddings.
It is defined as follows:

\begin{equation}
\label{eq:attention}
\alpha(\mathbf{A}|\mathbf{x},\mathbf{e}_1,\mathbf{e}_2) =
SoftMax\left(\sigma(
\mathbf{A}*
Q(\mathbf{x},\mathbf{e}_1)
K(\mathbf{x},\mathbf{e}_2)^T)
,\tau\right)
\end{equation}

where $*$ is an element-wise matrix multiplication, $\sigma$ is an activation function, and $\tau$ is the softmax temperature.
In equation \eqref{eq:attention}, the query $Q$ and key $K$ matrices are given through two FC layers:
\begin{align}
\label{eq:QK}
    K(\mathbf{x},\mathbf{e}_1) &= FC_{K}( \mathbf{x} + \mathbf{e_1} ),\\
    Q(\mathbf{x},\mathbf{e}_2) &= FC_{Q}( \mathbf{x} + \mathbf{e_2} ).
\end{align}

The query $Q(\mathbf{x},\mathbf{e}_2)$ and key $K(\mathbf{x},\mathbf{e}_2)$ matrices are functions of both the input $\mathbf{x}$ and node embeddings $\mathbf{e}_1$ and $\mathbf{e}_2$.
This formulation enables the SGT module to be adaptive to the unique and changing dynamics of each pair of sensors.

\subsubsection{Final FC and Loss}
The final FC layers take the aggregated output of the G-SWaN layers and act as a decoder.
In this sense, the G-SWaN layers act as a spatio-temporal encoder that produce compact representations of the input in contrast to multi-way tensor representations \cite{koniusz2021tensor}.
It outputs the forecast $\mathbf{h}$ for the next $F$ timesteps at all $N$ sensors. 

\subsubsection{Loss and Optimizer}
Following previous works, we use Mean Absolute Error (MAE) as the loss function and Adam with weight decay as the optimizer.

\subsubsection{Minor Improvements\label{sec:minor}}

Additionally, there are a number of minor modifications.
First, we adopt changes suggested by \cite{shleifer2019incrementally}:
(1) Adding a time-of-day feature and separate FC layer for embedding.
(2) Increasing the number of hidden channels from 32 to 40.
(3) Introducing a learning rate decay with a factor of 0.97 per epoch.
(4) Reducing the gradient clipping from 5 to 3 as additional residual connections are introduced around the GCN.
(5) Replacing missing data with the average of training data, instead of zero.

As well as the changes discussed above, a series of minor changes are introduced:
(1) Using Mish \cite{misra2019mish} as activation function.
(2) contrary to the finding in \cite{shleifer2019incrementally}, batch normalization is found to be useful and is implemented accordingly.

\section{Experiment}

\subsection{Dataset}
\label{sec:dataset}

\begin{table*}[htbp]
\centering
\small
\caption{Dataset description.}
\label{tab:dataset}
\begin{tabular}{@{}c|cc|cc|cc|cc@{}}
\toprule
\multirow{2}{*}{Dataset} & \multicolumn{2}{c|}{Spatial} & \multicolumn{2}{c|}{Temporal} & \multicolumn{2}{c|}{Value} & \multicolumn{2}{c}{Size} \\ \cmidrule(l){2-9}
 & Sensors & Edges & Timesteps & Range (duration in days) & Metric & Mean $\pm$ Std & Entry & Compressed (MB) \\ \midrule
METR-LA & 207 & 1,515 & 34,272 & 1 Mar 12 - 30 Jun 12 (121) & speed & 53.72$\pm$ 20.26 & 7,094,304 & 54 \\
PeMS-BAY & 325 & 2,694 & 52,116 & 1 Jan 17 - 30 Jun 17 (180) & speed & 62.61$\pm$ 9.59 & 16,937,700 & 130 \\
PeMS-D4 & 307 & 340 & 16,969 & 1 Jan 18 - 28 Feb 18 (58) & flow & 211.70$\pm$ 158.07 & 5,209,483 & 31 \\
PeMS-D8 & 170 & 277 & 17,833 & 1 Jul 16 - 31 Aug 16 (61) & flow & 230.68$\pm$ 146.22 & 3,031,610 & 18 \\ \bottomrule
\end{tabular}
\end{table*}

\begin{figure*}[htbp]
    \centering
    \includegraphics[width=0.99\textwidth]{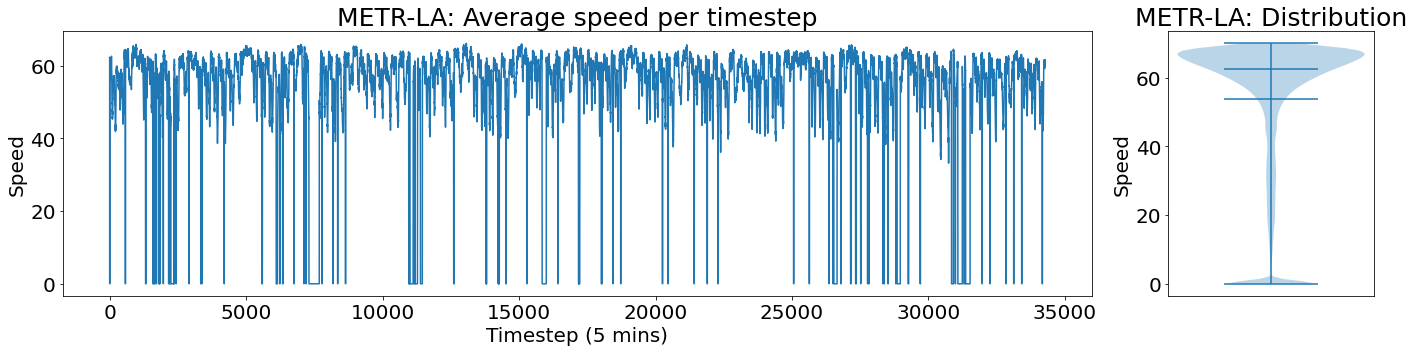}
    \includegraphics[width=0.99\textwidth]{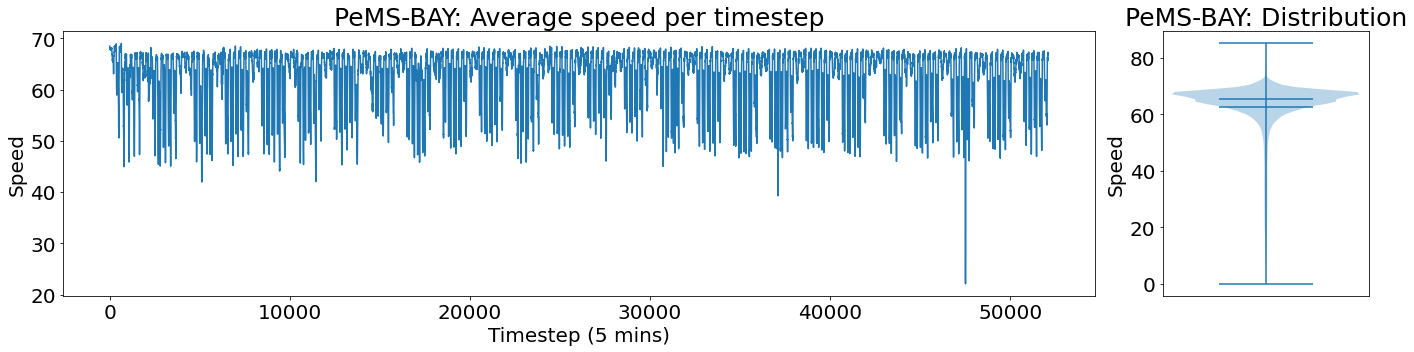}
    \includegraphics[width=0.99\textwidth]{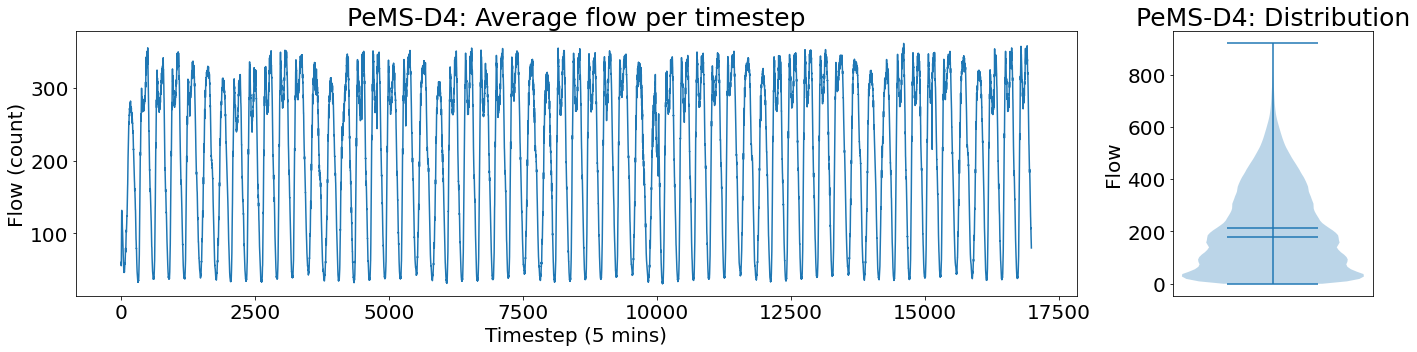}
    \includegraphics[width=0.99\textwidth]{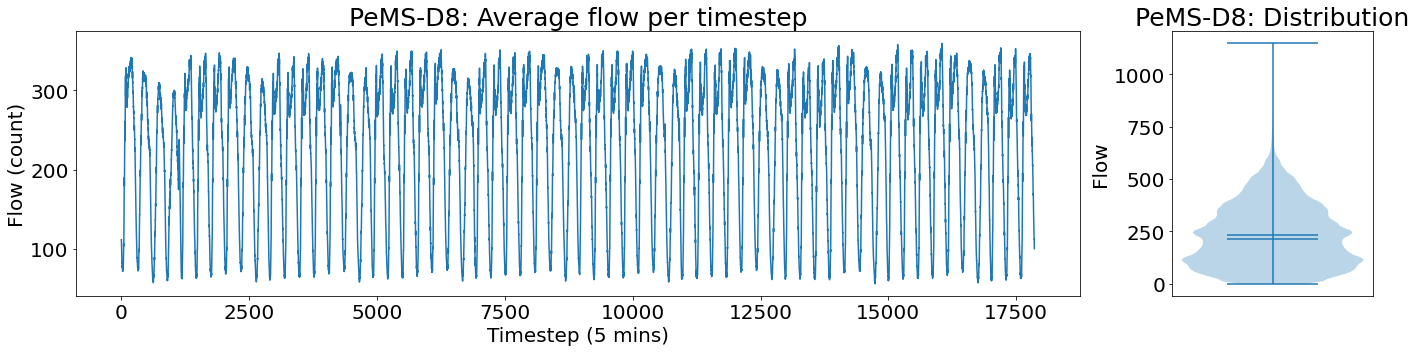}
    \caption{Visualization of the temporal dynamics and distribution of the datasets.}
    \label{fig:data}
\end{figure*}

We use four open, real-world, datasets that have been previously been used by other works as benchmarks \cite{li2017diffusion,yu2017stgcn,zhang2018gaan,wu2020graph,guo2019attention,song2020spatial,bai2020adaptive}.
The important statistics of these datasets are available in Table \ref{tab:dataset}.
All datasets record the relevant traffic metric every five minutes (one timestep) at various recording stations (sensors) in a road network.
The traffic metric recorded by METR-LA and PeMS-BAY is the average speed, while PeMS-D4 and PeMS-D8 is the flow (number of vehicles, also known as traffic volume).
They are all also accompanied with time-of-day information.
As can be seen in Figure \ref{fig:data}, the speed datasets have negative skews while the flow datasets have positive skews.
This is because most of the time, the road networks are not congested and vehicles are traveling at the speed limit. This is also shown through the slight negative correlation in the fundamental diagrams in Figure \ref{fig:fund_many}.
We selected these datasets because they are publicly available, and because they vary in important ways, as described below:
\begin{itemize}
    \item \textbf{METR-LA} \cite{jagadish2014big}. This dataset is collected from loop detectors in LA, USA county highway.
This is the most popular dataset for traffic forecasting.
% This dataset has the most missing values as shown in Figure \ref{fig:data};
    \item \textbf{PeMS-BAY} \cite{li2017diffusion}. This dataset is collected by California Transportation Agencies (CalTrans) Performance Measurement System (PeMS), also from loop detectors.
This is the second most popular dataset for traffic forecasting.
This is the largest dataset in terms of sensors, edges, timesteps, and entries among the four we are using.
    \item \textbf{PeMS-D4}. 
This dataset is also from CalTrans PeMS.
This, together with PeMS-D8, is the most popular dataset for traffic flow forecasting.
To the best of our knowledge, three works have also used both of these datasets~\cite{guo2019attention,song2020spatial,bai2020adaptive}.
It contains sensor data from freeways of district 4 in California, USA.
Temporally, this is the shortest dataset.
    \item \textbf{PeMS-D8}. 
This dataset is also from CalTrans PeMS.
All three works that used the PeMS-D4 dataset also used this dataset.
However, it is coming from district 8.
This is the smallest dataset in terms of sensors, edges, and number of entries.
\end{itemize}

All datasets are accompanied with an adjacency matrix based on the network distance between sensors.
To maintain consistency with previous works, we construct the adjacency matrix by passing the network distance between connected sensors through a Gaussian RBF:
\begin{equation}
    A_{i,j} = \exp\left(-\left(\frac{d(n_i,n_j)}{\sigma_d}\right)^2\right)
\end{equation}
where $A_{i,j}$ is the $i$-th row and $j$-th entry of adjacency matrix $A$,
$d(n_i,n_j)$ is the network distance between sensor $n_i$ and sensor $n_j$,
and $\sigma_d$ is the standard deviation of all the network distance between all immediately connected sensors.

\subsection{Experimental Setup}

We use the common experimental setup as typical in previous works.
One timestep consists of a 5-minute average speed for every sensor in the road network.
One datapoint of an input is made up of 12 consecutive timesteps (1 hour), and the output is the next 12 timesteps.

\subsubsection{Training, Validation, and Test Split}
In order to match the setting of previous works, the split ratio differs depending on the traffic metric.
For the datasets with the speed metric, we use 7:1:2 temporal split, following \cite{wu2020graph}.
For the datasets with the flow metric, we use 6:2:2 temporal split, following \cite{bai2020adaptive}.

\subsubsection{Feature Scaling}
We scale the speed channel of our input to achieve standard distribution as follows:
\begin{equation}
    x'= \frac{x-\mu_x}{\sigma_x}
\end{equation}
where $x$ is the raw input,
$x'$ is the input after scaling,
$\mu_x$ is the average of the raw input,
$\sigma_x$ is the standard deviation of the raw input.
This way, our input has zero mean and a standard deviation of one.
The time-of-day feature is scaled using MinMax scaler.

\subsubsection{Computing Infrastructure}
All of the models are implemented in Python 3.7.4 with PyTorch 1.5.1 (cuDNN 7.6.5 and CUDA 10.1). They are executed in a computing sensor with one NVIDIA Tesla V100 GPU card and Intel(R) Xeon(R) Platinum 8268 CPU @ 2.90GHz.

\subsubsection{Metrics}

To evaluate the prediction performance of the proposed methods, three widely used metrics are selected, namely Mean Average Error (MAE), Mean Average Percentage Error (MAPE), and Root Mean Square Error (RMSE). 
They are defined as follows:
\begin{align}
    MAE &= \frac{1}{N} \sum_{n=1}^{N} { \left| \mathbf{y}_n-\mathbf{h_n} \right| }, \\
    MAPE &= \frac{100\%}{N} \sum_{n=1}^{N}\left| \frac{\mathbf{y}_n-\mathbf{h}_n}{\mathbf{y}_n}\right| ,\\
    RMSE &= \sqrt{ \frac{1}{N}\sum_{n=1}^{N}(\mathbf{y}_n-\mathbf{h}_n)^2 } ,
\end{align}
where $\mathbf{y}$ is the actual label,
$\mathbf{h}$ is the predicted value,
and $N$ is the number of datapoints.

All three metrics are useful to compare the error profile of different models. MAE is the most interpretable since it is simply the average absolute error. RMSE is more sensitive to larger errors, thus when RMSE is significantly larger than MAE, it indicates that models make fewer mistakes, but those mistakes are larger in magnitude. This information is important for users of the traffic models. Finally, MAPE is unitless, making it useful for comparison between datasets.

Following the previous work, we report the performance at three forecasting horizons (15, 30, and 60 minutes) for the speed datasets (METR-LA and PeMS-BAY) \cite{li2017diffusion,wu2020graph}; whereas the average MAE, MAPE, and RMSE across the next 12 timesteps (corresponding to one hour) are reported for the flow datasets (PeMS-D4 and PeMS-D8) \cite{guo2019attention,song2020spatial,song2020spatial}.

\subsection{Baselines}
We compare G-SWaN with the performances of the following models:
(1) \textbf{HA} \cite{bai2020adaptive} Historical Average assumes daily periodicity and takes the average of the same time slot from the previous period;
(2) \textbf{ARIMA} \cite{hamed1995short} Auto-Regressive Integrated Moving Average;
(3) \textbf{VAR} \cite{zivot2006vector} Vector Auto Regressive;
(4) \textbf{SVR} \cite{li2021dynamic} Support Vector Regression;
(5) \textbf{FNN} \cite{li2021dynamic} Feed-forward Neural Network;
(6) \textbf{FC-LSTM} \cite{li2017diffusion} Fully Connected, Long Short-Term Memory network;
(7) \textbf{GRU-ED} \cite{cho2014learning} Gated Recurrent Unit Encoder Decoder;
(8) \textbf{DSANet} \cite{huang2019dsanet} Dual Self-Attention Network uses self-attention to capture spatial correlation;
(9) \textbf{DCRNN} \cite{li2017diffusion} Diffusion Convolution Recurrent Neural Network with graph convolution in the spatial block and GRU in the temporal block;
(10) \textbf{STGCN} \cite{yu2017stgcn} Spatio-Temporal Graph Convolutional Network, using spectral GCN;
(11) \textbf{Graph WaveNet} \cite{wu2020graph} uses spatial GCN \cite{hamilton2017inductive} in the spatial block and WaveNet \cite{oord2016wavenet} in their temporal block;
(12) \textbf{ST-MetaNet} \cite{pan2019urban} Spatio-temporal Meta-learning Network;
(13) \textbf{ASTGCN} \cite{guo2019attention} Attention-based Spatio-Temporal Graph Convolution Network uses both spatial and temporal attention;
(14) \textbf{STSGCN} \cite{song2020spatial} Spatio-Temporal Synchronous Graph Convolution Network captures spatio-temporal dynamics by adding temporal edges to their graph convolution;
(15) \textbf{AGCRN} \cite{bai2020adaptive} Adaptive Graph Convolutional Recurrent Network generates an adaptive spatial graph to complement the physical road network adjacency matrix;
(16) \textbf{GMAN} \cite{zheng2020gman} Graph Multi-Attention Network uses spatio-temporal and transform attention;
(17) \textbf{MTGNN} \cite{wu2020connecting} Multivariate Time-series Graph Neural Network uses meta-learning to learn the weights of the spatial and temporal module.

Besides for G-SWaN, the results in Table \ref{tab:metr} are based on the benchmark performed by \cite{li2021dynamic}, while for Table \ref{tab:pems}, it was performed by \cite{bai2020adaptive}. The exception is for Graph WaveNet in Table \ref{tab:pems}, which was reproduced by us.

\section{Results}

\subsection{Performance Comparison}

\begin{table*}[ht]
\centering
\caption{Performance comparison on speed metric using METR-LA and PeMS-BAY datasets.
Since all the metrics are error metrics, lower means better.
Prediction horizon = 15 / 30 / 60 minutes.
\textbf{Bold} means the best performance within the metric.
\underline{Underline} means the second best performance.
}
\label{tab:metr}
\normalsize
\begin{tabular}{@{}l|ccc|ccc@{}}
\toprule
\multicolumn{1}{c|}{(Metric: speed)}                & \multicolumn{3}{c|}{METR-LA}                                                & \multicolumn{3}{c}{PeMS-BAY}                      \\
Model                & MAE                          & RMSE                    & MAPE (\%)               & MAE                     & RMSE                    & MAPE (\%)               \\ \midrule
HA                   & 4.16                         & 7.80                    & 13.00                   & 2.88                    & 5.59                    & 6.80                    \\
ARIMA                & 3.99/5.15/6.90               & 8.21/10.45/13.23        & 9.60/12.70/17.40        & 1.62/2.33/3.38          & 3.30/4.76/6.50          & 3.50/5.40/8.30          \\
VAR                  & 4.42/5.41/6.52               & 7.89/9.13/10.11         & 10.20/12.70/15.80       & 1.74/2.32/2.93          & 3.16/4.25/5.44          & 3.60/5.00/6.50          \\
SVR                  & 3.99/5.05/6.72               & 8.45/19.87/13.76        & 9.30/12.10/16.7         & 1.85/2.48/3.28          & 3.59/5.18/7.08          & 3.80/5.50/8.00          \\
FNN                  & 3.99/4.23/4.49               & 7.94/8.17/8.69          & 9.90/12.90/14.00        & 2.20/2.30/2.46          & 4.42/4.63/4.98          & 5.19/5.43/5.89          \\
FC-LSTM              & 3.44/3.77/4.37               & 6.30/7.23/8.69          & 9.60/10.90/13.20        & 2.05/2.20/2.37          & 4.19/4.55/4.96          & 3.80/5.20/5.70          \\
DCRNN                & 2.77/3.15/3.60               & 5.38/6.45/7.60          & 7.30/8.80/10.50         & 1.38/1.74/2.07          & 2.95/3.97/4.74          & 2.90/3.90/4.90          \\
STGCN                & 2.88/3.47/4.59               & 5.74/7.24/9.40          & 7.62/9.57/12.70         & 1.36/1.81/2.49          & 2.96/4.27/5.69          & 2.90/4.17/5.79          \\
Graph WaveNet        & \underline{2.69}/3.07/3.53   & \underline{5.15}/6.22/7.37          & 6.90/8.37/10.01         & \textbf{1.30}/\underline{1.63}/1.95 & \underline{2.74}/\underline{3.70}/4.52          & \underline{2.73}/\underline{3.67}/4.63          \\
ST-MetaNet           & \underline{2.69}/3.10/3.69   & 5.17/6.28/7.52          & 6.91/8.57/10.63         & 1.36/1.76/2.20          & 2.90/4.02/5.06          & 2.82/4.00/5.45          \\
ASTGCN               & 4.86/5.43/6.51               & 9.27/10.61/12.52        & 9.21/10.13/11.64        & 1.52/2.01/2.61          & 3.13/4.27/5.42          & 3.22/4.48/6.00          \\
STSGCN               & 3.31/4.13/5.06               & 7.62/9.77/11.66         & 8.06/10.29/12.91        & 1.44/1.83/2.26          & 3.01/4.18/5.21          & 3.04/4.17/5.40          \\
AGCRN                & 2.87/3.23/3.62               & 5.58/6.58/7.51          & 7.70/9.00/10.38         & 1.37/1.69/1.96          & 2.87/3.85/4.54          & 2.94/3.87/4.64          \\
GMAN                 & 2.80/3.12/\textbf{3.44}      & 5.55/6.49/7.35          & 7.41/8.73/10.07         & 1.34/\underline{1.63}/\textbf{1.86} & 2.91/3.76/\textbf{4.32} & 2.86/3.68/\textbf{4.37} \\
MTGNN                & \underline{2.69}/\underline{3.05}/3.49   & 5.18/\underline{6.17}/\textbf{7.23} & \underline{6.86}/\underline{8.19}/\underline{9.87}          & \underline{1.32}/1.65/1.94          & 2.79/3.74/4.49          & 2.77/3.69/4.53          \\
\textbf{G-SWaN (ours)} & \textbf{2.65/3.02}/\underline{3.47}      & \textbf{5.05/6.12}/\underline{7.27} & \textbf{6.72/8.13/9.86} & \textbf{1.30/1.61}/\underline{1.91} & \textbf{2.72/3.64}/\underline{4.37} & \textbf{2.69/3.62}/\underline{4.49} \\ \bottomrule
\end{tabular}
\end{table*}

Table \ref{tab:metr} and \ref{tab:pems} present the results of our G-SWaN and other methods. For each column, the best result is given in bold. Note that HA makes the same prediction regardless of the forecasting horizon, so we only have one value for each metric.

When the forecasting horizon is increasing, the prediction performance of each method becomes worse, which is as expected.
As can be seen from Table \ref{tab:metr} and \ref{tab:pems}, the proposed G-SWaN achieves the best performances in all datasets, across all metrics, with few exceptions.
In those exceptions, G-SWaN always comes in the second place.
Overall, these results demonstrate the superior performance of our G-SWaN.

\begin{table}[htbp]
\centering
\caption{Performance comparison on flow metric using PeMS-D4 and PeMS-D8 datasets. 
Since all the metrics are error metrics, lower means better.
\textbf{Bold} means the best performance within the metric.
\underline{Underline} means the second best performance.
}
\label{tab:pems}
\addtolength{\tabcolsep}{-0.5ex}
\begin{tabular}{@{}l|ccc|ccc@{}}
\toprule
\multicolumn{1}{c|}{(Metric: flow)} & \multicolumn{3}{c|}{PeMS-D4}                     & \multicolumn{3}{c}{PeMS-D8}                    \\
Model                               & MAE            & RMSE           & MAPE       & MAE            & RMSE           & MAPE     \\ \midrule
HA                                  & 38.03          & 59.24          & 27.88          & 34.86          & 52.04          & 24.07         \\
VAR                                 & 24.54          & 38.61          & 17.24          & 19.19          & 29.81          & 13.10         \\
GRU-ED                              & 23.68          & 39.27          & 16.44          & 22.00          & 36.23          & 13.33         \\
DSANet                              & 22.79          & 35.77          & 16.03          & 17.14          & 26.96          & 11.32         \\
DCRNN                               & 21.22          & 33.44          & 14.17          & 16.82          & 26.36          & 10.92         \\
STGCN                               & 21.16          & 34.89          & 13.83          & 17.50          & 27.09          & 11.29         \\
Graph WaveNet                       & 28.98          & 42.08          & 30.80          & 20.52          & 30.04          & 16.20         \\
ASTGCN                              & 22.93          & 35.22          & 16.56          & 18.25          & 28.06          & 11.64         \\
STSGCN                              & 21.19          & 33.65          & 13.90          & 17.13          & 26.86          & 10.96         \\
AGCRN                               & \underline{19.83}          & \underline{32.26}          & \underline{12.97}          & \underline{15.95}          & \underline{25.22}          & \underline{10.09} \\
\textbf{G-SWaN (ours)}                & \textbf{18.48} & \textbf{30.51} & \textbf{12.59} & \textbf{14.05} & \textbf{23.00} & \textbf{9.08} \\
\bottomrule
\end{tabular}
\end{table}

Since MAPE is unitless, it provides a useful metric to compare different datasets.
The results show that some datasets are easier than others (i.e. achieving lower error), because the datasets have different variability in the first place.
Broadly speaking, the speed datasets seem to be easier than the flow datasets.
PeMS-BAY is the easiest dataset, where all models seem to have low MAPE, including the non-learning baselines HA and ARIMA.
Since there is no learning with HA and ARIMA, the reason cannot be that PeMS-BAY has more datapoints to learn from, but rather that it has low variability, as shown with low standard deviation in Table \ref{tab:dataset}, even as a proportion to the mean value $15.32\%$.
Using this ratio between standard deviation to mean, we could rank the difficulty of the datasets.
From easiest to hardest, the subsequent ranking is, METR-LA ($37.7\%$), PeMS-D8 ($63.39\%$), and finally PeMS-D4 ($74.67\%$) as the most difficult.
Moreover, the difference between the spread of different datasets can also be observed visually in Figure \ref{fig:data}.
The speed datasets usually hover around the speed limit, while the value of the flow datasets cycles more drastically throughout the day, resulting in a wider distribution, as seen from the violin plots.

This method of ranking is mostly correct when comparing the MAPE across the datasets, except that PeMS-D8 is more difficult than PeMS-D4, despite the lower standard deviation, both in absolute value and proportion.
Again, this could be explained, not by the lack of training data but by the lack of periodic behavior.
This can be shown by the disparity of the MAPE error for the non-learning model HA, which is simply the average value from the day before.

G-SWaN performance gain seems to be relative to the difficulty of the dataset.
For example, at PeMS-BAY, since it is an easier datasets \cite{li2021dynamic} and all the other models are already performing relatively well, our improvements are marginal.
At 15 minutes forecasting horizon, G-SWaN only achieves the same MAE with Graph WaveNet,
while at 60 minutes horizon, G-SWaN was outperformed by GMAN.

With the relative difficulty of the datasets discussed, we can compare the models across datasets.
There are nine models that are implemented across all datasets: HA, VAR, DCRNN, STGCN, Graph WaveNet, ASTGCN, STSGCN, AGCRN, and G-SWaN.
As expected, deep learning models outperformed their classical counterparts.
Simply adding an attention mechanism does not guarantee improvement, however as shown through the performance of ASTGCN, as opposed to GMAN and G-SWaN.

Both G-SWaN and AGCRN employ a data adaptive adjacency matrix.
AGCRN used matrix factorization to achieve this, while G-SWaN used a self-attention mechanism.
The superior performance of G-SWaN corroborates the literature regarding the importance of self-attention mechanisms when introducing data adaptive components.

\begin{table}[htbp]
\centering
\caption{Ablation study on PeMS-D8.
\textbf{Bold} means the worst performance, showing the importance of the missing component.
\underline{Underline} means the second worst.
}
\label{tab:ablation}
\normalsize
\begin{tabular}{@{}l|ccc@{}}
\toprule
Model                                         & MAE   & MAPE (\%) & RMSE \\ \midrule \midrule
\textbf{G-SWaN}            & 14.05 & 9.08 & 23.00     \\
\midrule
w/o spatial occlusion    & 14.12 & 9.14 & 23.10     \\
w/o temporal permutation & 14.21 & 9.18 & \underline{23.15}     \\
w/o uniform noise        & 14.11 & \underline{9.22} & \underline{23.15}     \\
w/o node embeddings                 & \underline{14.29} & \underline{9.22} & 23.11     \\
Single head attention    & 14.21 & 9.14 & 23.05     \\
GCN w/o SGT              & \textbf{14.62} & \textbf{9.52} & \textbf{23.34}     \\ \bottomrule
\end{tabular}
\end{table}

\subsection{Ablation}

We performed ablation analysis on the PeMS-D8 dataset, as shown in Table \ref{tab:ablation}.
We used the same experimental setup for the PeMS-D8 dataset.
We named G-SWaN without different components as follows:
\begin{enumerate}
    \item \textbf{w/o spatial occlusion:} G-SWaN without the spatial occlusion augmentation as described in section \ref{sec:aug};
    \item \textbf{w/o temporal permutation:} G-SWaN without the temporal permutation augmentation as described in section \ref{sec:aug};
    \item \textbf{w/o uniform noise:} G-SWaN without the uniform noise augmentation as described in section \ref{sec:aug};
    \item \textbf{w/o node embeddings:} G-SWaN without having the node embeddings fused with the keys and queries matrices in SGT. See Equation \ref{eq:QK};
    \item \textbf{Single head attention:} G-SWaN where the SGT only had one attention head ($H=1$);
    \item \textbf{GCN w/o SGT:} G-SWaN where we replaced the SGT with the GCN described in \cite{wu2020graph}.
\end{enumerate}

The ablation analysis results show that each of the components of G-SWaN is effective.
In particular, the results show that the worst is the variant without SGT (in \textbf{bold}), the main contribution of this paper.
Moreover, removing the node embeddings from the SGT module also degrades the MAE and MAPE performances significantly, resulting in the second worse version (\underline{underlined}).
This agrees with our hypothesis that SGT and integrating node embeddings in the self-attention mechanism improved the performances by capturing the individual sensor dynamics, the pair dynamics, and the evolution of the dynamics through time.

\subsection{Node Embedding Analysis}

\begin{figure}[htbp]
\centering
\includegraphics[width=.7\columnwidth]{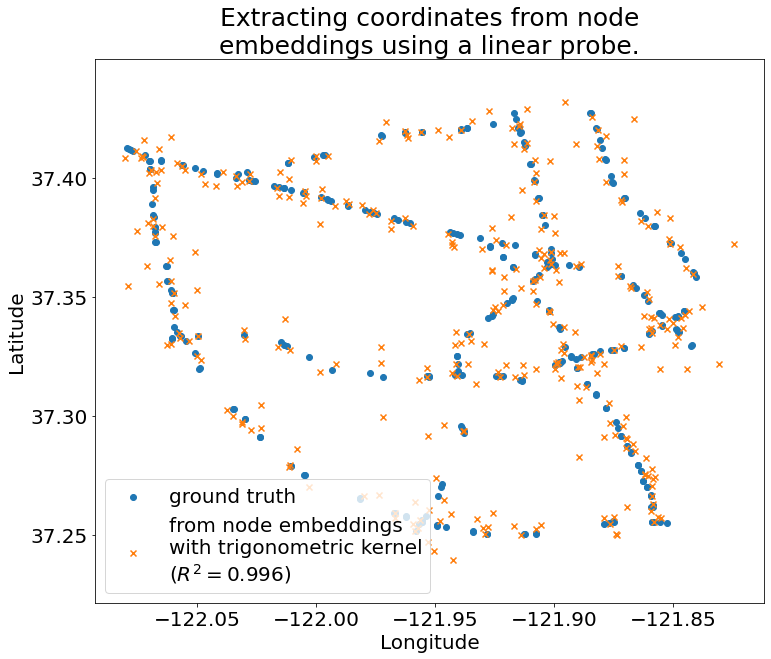}
\caption{Recovering sensor coordinates from node embeddings using a linear probe with trigonometric kernels.}
\label{fig:probe}
\end{figure}

To further investigate the node embeddings, we wanted to see if the node embeddings managed to learn spatial information that was not part of the training data.
For this purpose, we tried to find any linear and trigonometric relationships between the node embeddings and the spatial coordinates of each sensor (longitude and latitude).
We used Coefficient of determination ($R^2$) as a metric.
This was implemented in the dataset with the most sensors, PeMS-BAY.

First, we performed linear regression on each sensor, using a concatenation of source $\mathbf{e}_1$ and target $\mathbf{e}_2$ node embeddings as the feature vector, and the longitude and latitude as the labels.
We obtained $R^2 = 0.401$, showing a non-trivial correlation between the node embeddings with the coordinates of the sensors.
This confirms that the node embeddings do contain spatial information.

Since the node embeddings were learned as a part of a non-linear deep learning method, there was no guarantee that linear probing was the best tool to analyze the learned features.
To address this, we used three trigonometric functions as kernels for the node embeddings: $sin(\cdot)$, $cos(\cdot)$, and $tan(\cdot)$.
Using linear regression with the kernels, we obtained $R^2 = 0.996$.
To visually evaluate this, we plotted the coordinates' value based on the linear probing of the node embeddings with trigonometric kernels as shown in Figure \ref{fig:probe}.
The figure shows that all the orange crosses are located close to the ground truth (blue circles).
This again confirms that node embeddings learned spatial information from the data.

% \begin{figure}[htbp]
%     \centering
%     \subfigure[Recovering sensor coordinates from node embeddings using a linear probe with trigonometric kernels. \label{fig:probe}]{
%         \includegraphics[width=.25\columnwidth]{coordinates_from_linear_probe_e03_enb_inspector3.png}
%     }
%     \hfill
%     \subfigure[Qualitative comparison between physical and adaptive adjacency matrix in METR-LA dataset. The line transparency is proportional to the edge weight. \label{fig:adj}]{
%         \includegraphics[width=.72\columnwidth]{conn_SIDEbySIDE.png}
%     }
%     \caption{Spatial analysis}
% \end{figure}

\subsection{Adaptive Adjacency Matrix Analysis}

\begin{figure}[htbp]
\includegraphics[width=.6\columnwidth]{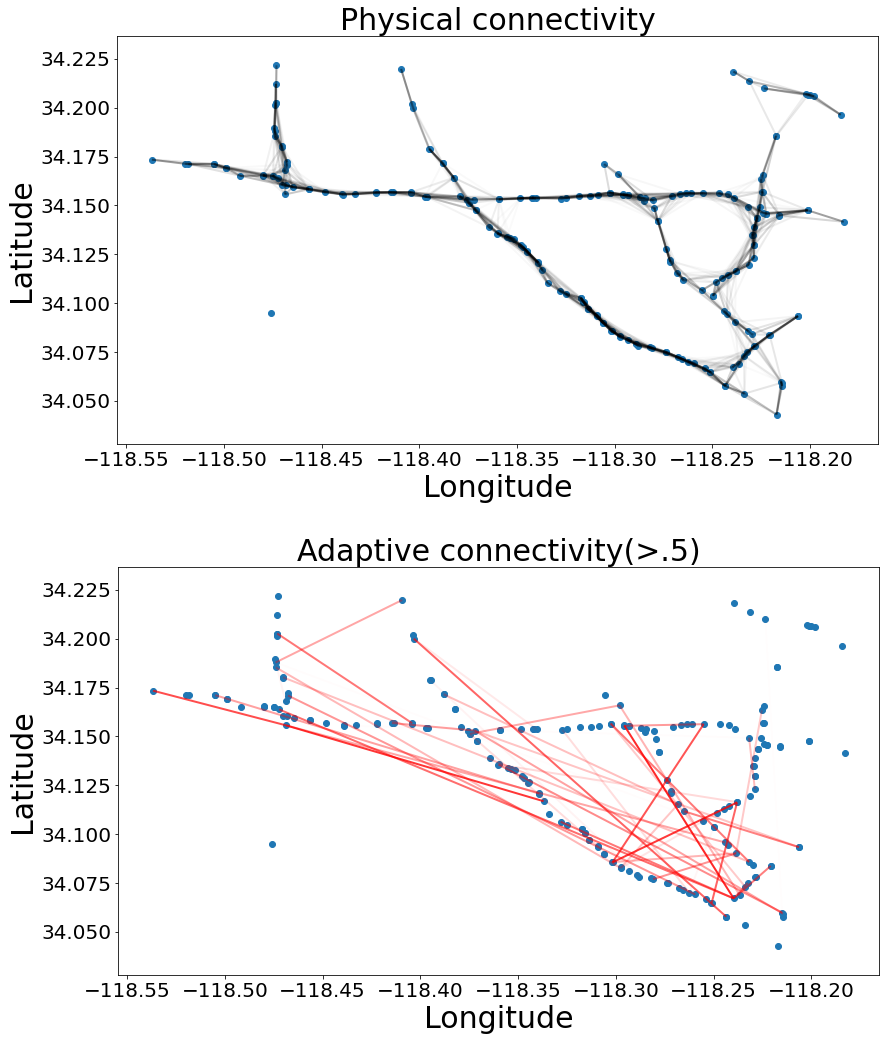}
\caption{Qualitative comparison between physical and adaptive adjacency matrix in METR-LA dataset.
The line transparency is proportional to the edge weight.
}
\label{fig:adj}
\end{figure}

Adaptive adjacency matrix $\mathbf{A}_{adp}$ is a learned adjacency matrix described in section \ref{sec:sgt}.
It is also constructed based on node embeddings.

Its purpose is to capture interesting connections between sensors that are not due to physical connections in the road network.
Intuitively, therefore, it should have minimal overlap with the adjacency matrix from the physical road network.
This can be qualitatively shown in Figure \ref{fig:adj}.
The figure shows that the physical edges (black lines) connect nearby sensors, while the adaptive adjacency (red lines) connects the sensors that are physically far away.
To further test this empirically, we flattened the adjacency matrices and calculated the cosine similarity between the physical and adaptive ones.
The similarity is 0.0226, which shows that they are widely different.

\section{Conclusion}
In this paper, we show that each sensor in a road network has a unique traffic dynamic.
Moreover, each pair of sensors also has a unique dynamic that displays periodic behaviors.
To capture these phenomena, we present G-SWaN, a novel traffic forecasting architecture.
This behavior can be handled by SGT, which is proposed to replace GCN, a widely used model for traffic forecasting, since SGT is more general and able to mask the adjacency matrix adaptively to the data.
In order to make the SGT adaptive to the sensor location as well, it incorporates node embeddings that adapt the self-attention mechanism with spatial information and sensor-unique dynamics for every source-target sensor pair.
Our experiments on four open, real-world datasets show that G-SWaN achieved state-of-the-art performance.
Finally, through recovering the co-ordinates, we show that the node embeddings learned meaningful spatial information.
These findings should inform future traffic models regarding the importance of capturing sensors and sensor pairs unique dynamics.

\bibliographystyle{ACM-Reference-Format}
\bibliography{1bib}

\begin{acks}
This research is supported by Australian Research Council (ARC) Discovery Project DP190101485.
We would like to also acknowledge the support of the Investigative Analytics team (Data61/CSIRO). We would also like to acknowledge the support of Cisco's National Industry Innovation Network (NIIN) Research Chair Program.
\end{acks}

\end{document}